\documentclass[letterpaper, 10 pt, conference]{ieeeconf}  

\IEEEoverridecommandlockouts                              

\usepackage{hyperref}
\hypersetup{
    colorlinks=true,
    linkcolor=blue,
    filecolor=magenta,      
    urlcolor=cyan,
    pdftitle={Overleaf Example},
    pdfpagemode=FullScreen,
    }
\usepackage{gensymb}
\usepackage{times}
\usepackage{epsfig}
\usepackage{graphicx}
\usepackage{amsmath,amssymb,amsfonts}
\usepackage{booktabs}
\usepackage[noend]{algpseudocode}
\usepackage{balance}
\usepackage[svgnames]{xcolor}
\usepackage{url}
\usepackage{authblk}
\usepackage{multirow}
\newcommand{\eg}{\emph{e.g.},}
\newcommand{\ie}{\emph{i.e.},}
\newcommand{\etal}{\emph{et~al.}}
\def\secref#1{Section~\ref{#1}}
\def\figref#1{Figure~\ref{#1}}
\def\tabref#1{Table~\ref{#1}}
\def\eqref#1{Equation~(\ref{#1})}

\usepackage{arydshln}
\usepackage{cite}
\usepackage{nicematrix}
\usepackage{subcaption}
\usepackage{graphbox}
\usepackage{soul}
\usepackage{svg}
\usepackage{color, colortbl}
\usepackage{tabularx}

\usepackage{booktabs}
\usepackage[linesnumbered,ruled]{algorithm2e}
\usepackage[flushleft]{threeparttable}
\usepackage{lipsum}
\usepackage{amssymb}\usepackage{pifont}\newcommand{\cmark}{\ding{51}}\newcommand{\xmark}{\ding{55}}\newcommand\numberthis[1][]{\refstepcounter{equation}\ifx#1\empty\else\label{eq:#1}\fi \tag{\theequation}}

\definecolor{darkgreen}{rgb}{0,0.5,0}
\definecolor{salmon}{rgb}{1.0, 0.55, 0.41}

\newcommand{\datashort}{\textit{Forest I}}
\newcommand{\datalong}{\textit{Forest II}}
\newcommand{\datavenman}{\textit{Forest III}}
\newcommand{\xbf}{\boldsymbol{x}}

\newcommand{\degr}{$^{\circ}$}

\makeatletter
\def\BState{\State\hskip-\ALG@thistlm}
\makeatother
\algdef{SE}[DOWHILE]{Do}{doWhile}{\algorithmicdo}[1]{\algorithmicwhile\ #1}

\graphicspath{ {figures/} }

\usepackage{arydshln}
\newcolumntype{x}[1]{>{\centering\arraybackslash\hspace{0pt}}p{#1}}
\newcolumntype{M}[1]{>{\centering\arraybackslash}m{#1}}
\newcolumntype{L}[1]{>{\raggedright\arraybackslash} m{#1} }

\usepackage{titlesec}

\begin{document}

\title{\LARGE \bf
{Online 6DoF Global Localisation in Forests using Semantically-Guided Re-Localisation and Cross-View Factor-Graph Optimisation }
}

\author{Lucas Carvalho de Lima$^{1,2}$, Ethan Griffiths$^{1,3}$, Maryam Haghighat$^3$, Simon Denman$^3$, \\ Clinton Fookes$^3$, Paulo Borges$^{2}$, Michael Brünig$^2$ and Milad Ramezani$^{1*}$ 

\thanks{
$^1$ CSIRO Robotics, DATA61, CSIRO, Brisbane, Australia. 
E-mails: {\tt\footnotesize \emph{
firstname.lastname
}@data61.csiro.au}}
\thanks{
$^{2}$ School of Information Technology and Electrical Engineering, The University of Queensland (UQ), Brisbane, Australia.
}
\thanks{
$^{3}$ School of Electrical Engineering and Robotics, Queensland University of Technology (QUT), Brisbane, Australia.}
\thanks{
$^*$Corresponding author
}
}

\maketitle

\renewcommand{\thefootnote}{\fnsymbol{footnote}}

\begin{abstract}

This paper presents FGLoc6D, a novel approach for robust global localisation and online 6DoF pose estimation of ground robots in forest environments by leveraging deep semantically-guided re-localisation and cross-view factor graph optimisation. The proposed method addresses the challenges of aligning aerial and ground data for pose estimation, which is crucial for accurate point-to-point navigation in GPS-degraded environments. By integrating information from both perspectives into a factor graph framework, our approach effectively estimates the robot's global position and orientation. Additionally, we enhance the repeatability of deep-learned keypoints for metric localisation in forests by incorporating a semantically-guided regression loss. This loss encourages greater attention to wooden structures,~\eg~tree trunks, which serve as stable and distinguishable features, thereby improving the consistency of keypoints and increasing the success rate of global registration, a process we refer to as re-localisation. The re-localisation module along with the factor-graph structure, populated by odometry and ground-to-aerial factors over time, allows global localisation under dense canopies. We validate the performance of our method through extensive experiments in three forest scenarios, demonstrating its global localisation capability and superiority over alternative state-of-the-art in terms of accuracy and robustness in these challenging environments.  
Experimental results show that our proposed method can achieve drift-free localisation with bounded positioning errors, ensuring reliable and safe robot navigation through dense forests.

\end{abstract}

\renewcommand*{\thefootnote}{\arabic{footnote}}

\section{Introduction}
\label{sec:intro}

Global localisation is a well-defined problem in field robotics; however it remains challenging to solve in forest environments. Current localisation solutions integrate onboard proprioceptive and exteroceptive measurements for pose estimation, commonly referred to as odometry~\cite{aftatah2016fusion, legoloam18, dusha2012error}, or Simultaneous Localisation and Mapping (SLAM) methods~\cite{Aguiar2020}, which leverage loop closures for improved accuracy\----albeit requiring the robot to revisit previously seen locations over time.

LiDAR-based SLAM systems build high-resolution maps that are valuable for forest inventory applications~\cite{Nie2021}. However, in long-range point-to-point navigation tasks \----our primary focus\---- LiDAR-SLAM methods suffer from drift over long runs in open-loop routes.
Even when integrated with GPS, SLAM can provide geo-referenced localisation~\cite{shan2020lio, Kukko2017}, but the resulting position estimates remain subject to GPS inaccuracies as GPS signals are often obstructed or degraded by foliage.

\begin{figure}
    \centering
    \includegraphics[width=.95\linewidth]{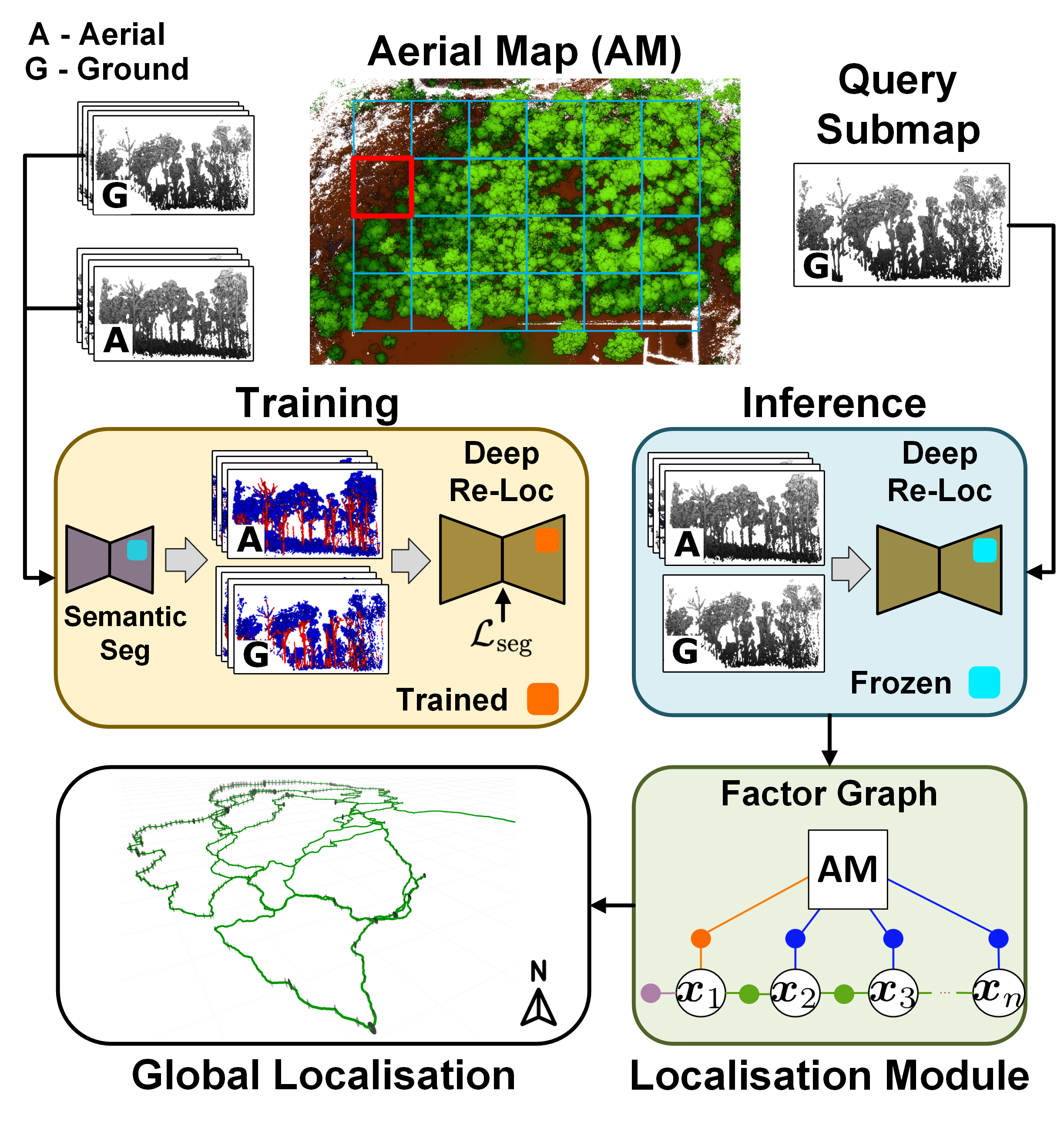}
    \caption{\small{Our method, \emph{FGLoc6D}, incorporates a Semantically-Guided Re-Localisation method for cross-view place retrieval and robust keypoint registration. Following re-localisation, we employ a factor-graph optimisation framework that integrates LiDAR-inertial odometry and ground-to-aerial registration factors, obtaining accurate global 6DoF pose estimation within the aerial reference map.}}
    \label{fig:ellipsoids}
    \vspace{-6mm}
\end{figure}
The primary challenge we address is to achieve reliable global localisation for ground robots navigating dense forest environments, ensuring they can reach target locations over long-range, open-loop trajectories. Inspired by our previous work~\cite{de2023air}, we introduce a novel aerial-ground collaboration method, dubbed \emph{FGLoc6D}, that significantly enhances localisation accuracy and robustness. Our approach leverages deep-learned re-localisation to efficiently position ground robots within an aerial LiDAR map (\figref{fig:ellipsoids}), diminishing the reliance on the assumption that the robot starts from a known initial position\----a key distinction from our previous work~\cite{de2023air}. 

LiDAR Place Recognition (PR) and cross-view data association between aerial and ground views present unique challenges in forest areas due to highly irregular and unstructured features (\eg~vegetation), often leading to degeneracy. To address this, we propose a semantically-guided loss function that prioritises features on the wooden structure of trees,~\eg~tree trunks, during training, improving the stability of point features for ground-to-aerial registrations. Once re-localisation succeeds, we locally register ground submaps with their aerial correspondences to estimate 6DoF robot poses using factor-graph optimisation and incorporation of uncertainties derived from data association. This differs from our earlier solution, which relies on Monte Carlo Localisation (MCL) and estimates 3DoF robot poses.

The main contributions are summarised as follows: 
\begin{itemize} 
\item We present an effective online 6DoF geo-localisation system, FGLoc6D, for ground robots operating in GPS-degraded forest environments. 
\item We introduce a semantically-guided method to increase repeatability of keypoints for robust global localisation.
\item We formulate ground-to-aerial correspondences as unary factors within factor-graph optimisation integrated with deep-learned re-localisation. 
\item We validate the proposed pipeline through extensive experiments conducted under canopies, demonstrating its robustness and efficacy. 
\end{itemize}

 \section{Related Work}
\label{sec:relwork}
This section reviews current localisation pipelines using aerial data and then discusses re-localisation research.
\subsection{Localisation Using Aerial Data}
Ground-to-aerial robot localisation has been addressed either by aligning local range data with maps of building structures and edges derived from satellite imagery~\cite{Kim2019, Tang2021} or learning to associate cross-view visual features~\cite{Wang_2024_CVPR, fervers2023uncertainty}. Although promising, these methods are not well-suited for environments with complex geometries, such as forests.

Tackling ground robot localisation in natural environments, Vandapel~\etal~\cite{Vandapel2006} match terrain surface features extracted from cross-view LiDAR data; however, this approach struggles in featureless terrains. Although alternative solutions exist, fusing traditional features obtained from ground and satellite images~\cite{Viswanathan2014} or low and high-altitude drone images~\cite{Shalev2020}, vision-based methods falter in dense forests due to canopy occlusions. Hussein \etal~\cite{Hussein2013} use Iterative Closest Point (ICP)~\cite{besl1992method} to align 3D ground lidar data of tree trunks with overhead crown maps; however, their approach is sensitive to errors in crown edge delineation, leading to potential localisation inaccuracies.

Carvalho de Lima~\etal~\cite{de2023air} localise ground robots within an aerial map by extracting cross-view invariant features from trees and using a particle filter to score and match hypotheses. However, their approach estimates only 3DoF poses. A tightly coupled LiDAR-inertial localisation on prior ground-view maps using factor graphs is presented in \cite{Kenji24}. While similar to our approach, it is limited to urban areas and 2D occupancy submaps for re-localisation, hence unsuitable for forest areas. Additionally, our method leverages map factors from aerial data for 6DoF pose estimation and a deep re-localisation module tested in complex forest scenarios. 

\subsection{LiDAR-based Re-localisation}

Re-localisation with LiDAR determines a robot's pose by matching current scans to a 3D prior map, comprising LiDAR Place Recognition (PR) and relative transformation calculation. Conventional LiDAR PR methods,~\eg~\cite{kim2018scan, salti2014shot, dube2017segmatch}, encode point clouds into global or local descriptors; however, they struggle in unstructured environments,~\eg~Scan Context~\cite{kim2018scan} as reported in~\cite{knights2022wild}.

Deep-learned approaches, on the other hand, use neural networks to extract features, which are either used directly~\cite{dube2018segmap, tinchev2019learning} or aggregated into global descriptors~\cite{radenovic2018fine, arandjelovic2016netvlad} for LiDAR PR. Methods such as EgoNN~\cite{komorowski2021egonn} and LCDNet~\cite{cattaneo2022lcdnet} estimate the relative pose between two point clouds upon success in LiDAR PR by associating local features using RANSAC or Optimal Transport (OT). To increase re-localisation robustness, authors have proposed ranking methods of top-k candidates~\cite{vidanapathirana2023spectral}, deep-learned cross-modal modules~\cite{ramezani2023deep}, and incorporation of semantic descriptors~\cite{Jin22}. Instead, we rely on geometry-based global descriptors for PR and introduce a semantically-guided loss function to weight keypoints placement on tree trunks, hence improving their repeatability and robustness during metric localisation. 

 \section{Method}
\label{sec:method}

\subsection{System Overview}

We aim to localise a ground robot against a prior aerial lidar map $\mathcal{M}$ that we split into $N_A$ submaps $\{\mathcal{P}_A^{[i]} \in \mathbb{R}^3\}_{i=1}^{N_A}$, each of which covers the same area as the ground lidar point clouds $\{\mathcal{P}_G^{[j]} \in \mathbb{R}^3\}_{j=1}^{N_G}$, where $N_G$ indicates number of ground submaps generated over time.

We formulate the problem as a bipartite graph, $\mathcal{G}=(\mathcal{U},\mathcal{V},\mathcal{E})$, consisting of variable nodes $\mathcal{V}$, in our case robot poses, and factor nodes $\mathcal{U}$, representing the conditional probability between connected variable nodes. Edges $\mathcal{E}$ in the graph connect factor nodes to variable nodes, signifying the dependency of the factors on the variables.

\begin{figure*}
    \centering
 \includegraphics[width=1.91\columnwidth]{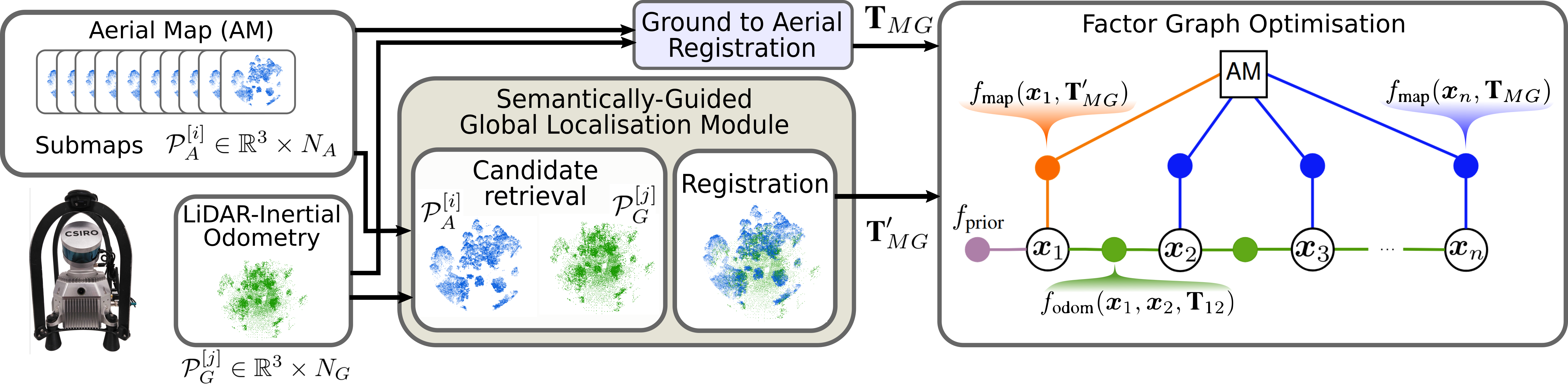}
    \caption{\small{Block diagram of our proposed pipeline. \emph{FGLoc6D} is composed of a semantically-guided deep module to re-localise the ground robot against an aerial map. Once re-localisation is done, the method estimates 6DoF robot poses using a fixed-lag smoothing factor graph structure. The factor graph is populated with a prior factor, odometry factors obtained from LiDAR-inertial odometry, and unary factors representing ground and aerial submap registrations.}}
    \label{fig:system_overview}
    \vspace{-5mm}
\end{figure*}

In our factor graph (\secref{sec:ag_factor_graph}), we define two types of factors: binary and unary. Binary factors are the odometry factors computed from a LiDAR-inertial odometry system. Unary factors are point cloud registrations between ground and aerial submaps. 

To solve global localisation, upon generation of a new submap,~\ie~a query point cloud $\mathcal{P}_G^{[q]}$, we run a deep LiDAR PR network, described in~\secref{sec:egonn}. This compares $\mathcal{P}_G^{[q]}$ with all the submaps $\mathcal{P}_A^{[i]}$ of the prior map to find the top candidate, $\mathcal{P}_A^{[t]}$, using a similarity metric. Initial relative pose $\mathbf{T}'_{MG}\in SE(3)$ between submaps $\mathcal{P}_G^{[q]}$ and $\mathcal{P}_A^{[t]}$ is further estimated using corresponding keypoints (\secref{sec:egonn}). To increase robustness, we apply RANSAC to associate keypoint local features and obtain the initial registration. Final 6DoF registration is later computed using Generalised ICP (GICP)~\cite{segal2009generalized}. For verification, we compute a fitness score out of ICP registration to decide the success of re-localisation. 

The factor derived from our deep-learned re-localisation module, in conjunction with other factors, is used in factor-graph optimisation to estimate 6DoF robot poses with respect to the aerial map.~\figref{fig:system_overview} depicts our proposed pipeline.

\subsection{Semantically-Guided Deep Re-Localisation}
\label{sec:egonn}
To localise the robot against prior map $\mathcal{M}$, our re-localisation module builds on EgoNN~\cite{komorowski2021egonn}, employing a light 3D CNN to train a global descriptor $d_{\mathcal{G}}\in \mathbb{R}^{256}$ and multiple local embeddings $\{d_{\mathcal{L}_t}\}_{t=1}^M\in \mathbb{R}^{128}$ with keypoint coordinates $\mathcal{Q} \in \mathbb{R}^{M\times3}$, where $t$ indexes the $M$ keypoints detected by USIP~\cite{li2019usip} in each submap. The global descriptor is derived from pooling feature maps $\mathcal{F}_{\mathcal{G}}\in \mathbb{R}^{K\times 128}$ using GeM~\cite{radenovic2018fine}, with $K$ representing the number of local features. Keypoint descriptors are computed in the local head by processing the local feature map $\mathcal{F}_{\mathcal{L}}\in \mathbb{R}^{M\times 64}$, with local embeddings $d_{\mathcal{L}_t}$ generated by a two-layer Multi-Layer Perceptron (MLP) followed by $L_2$ normalisation. Global descriptors are used for place recognition, whilst local descriptors support metric localisation.

\begin{figure}[t]
  \centering
  \begin{subfigure}{0.49\linewidth}
    \setlength{\fboxsep}{1pt}
    \fcolorbox{green}{green}{\includegraphics[trim={0 1.0cm 0 .3cm}, clip, width=.95\linewidth]{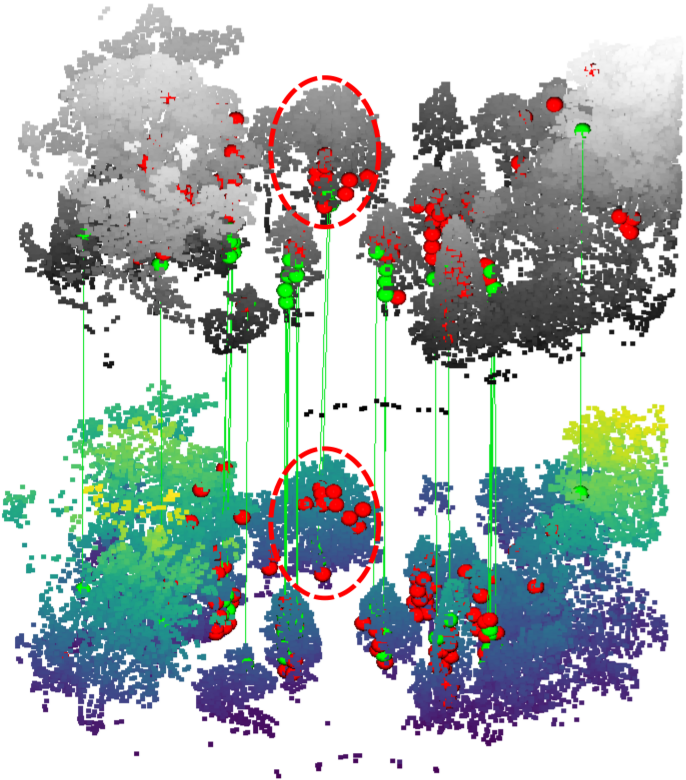}}
    \caption{EgoNN (localisation success)}
    \label{fig:egonn_base_kp_success}
  \end{subfigure}
  \hfill
  \begin{subfigure}{0.49\linewidth}
    \setlength{\fboxsep}{1pt}
    \fcolorbox{green}{green}{\includegraphics[trim={0 1.0cm 0 .3cm}, clip, width=.95\linewidth]{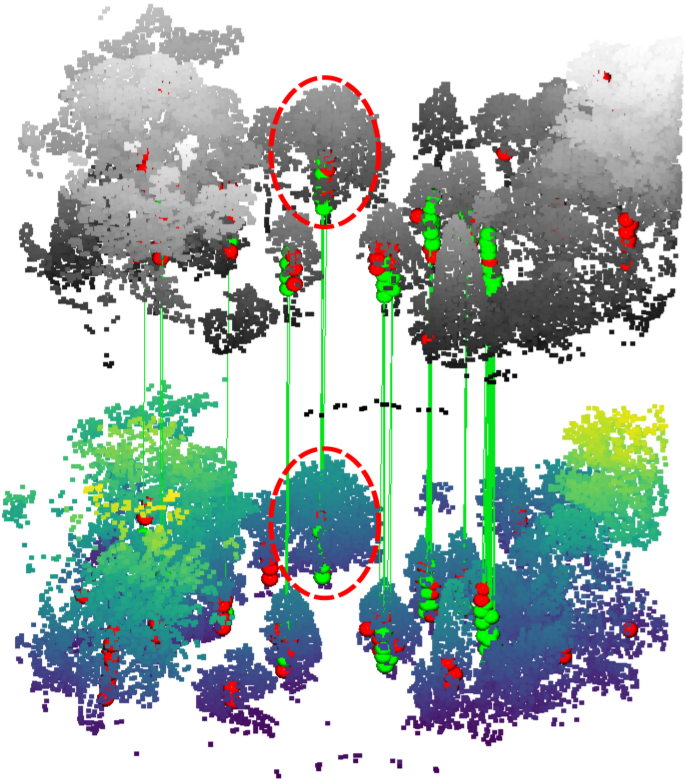}}
    \caption{\textbf{Ours} (localisation success)}
    \label{fig:egonn_semantic_kp_success}
  \end{subfigure}\\
\vspace{1mm}
  \begin{subfigure}{0.49\linewidth}
    \setlength{\fboxsep}{1pt}
    \fcolorbox{red}{red}{\includegraphics[trim={0 .4cm 0 .4cm}, clip, width=.97\linewidth]{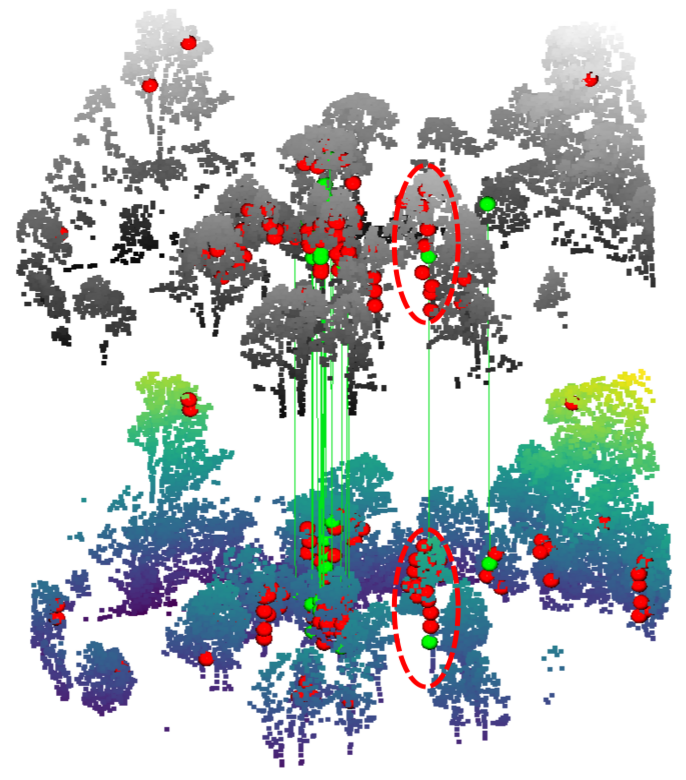}}
    \caption{EgoNN (localisation failure)}
    \label{fig:egonn_base_kp_failure}
  \end{subfigure}
  \hfill
\begin{subfigure}{0.49\linewidth}
    \setlength{\fboxsep}{1pt}
    \fcolorbox{green}{green}{\includegraphics[trim={0 .4cm 0 .4cm}, clip, width=.97\linewidth]{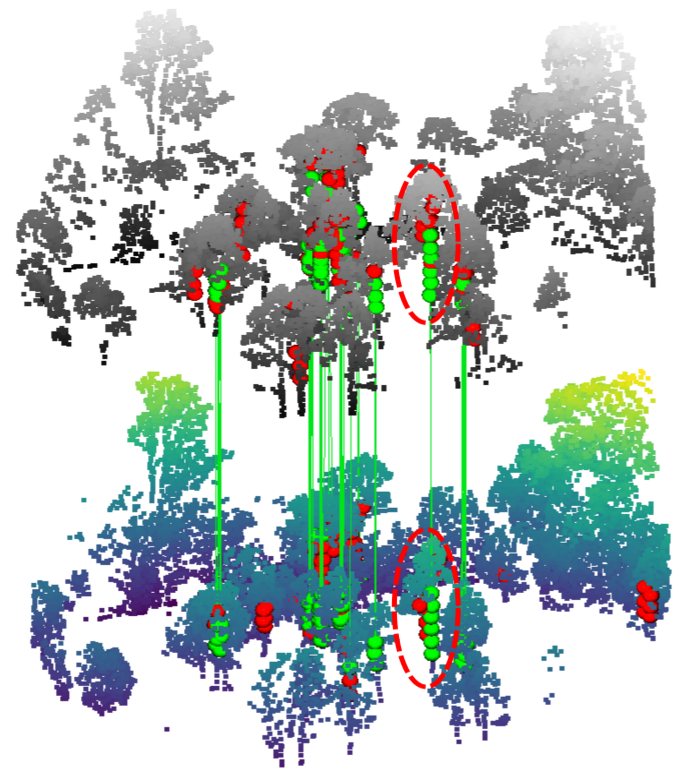}}
    \caption{\textbf{Ours} (localisation success)}
    \label{fig:egonn_semantic_kp_failure}
  \end{subfigure}
\caption{\small{EgoNN keypoint degeneracy in ground and aerial submap pairs. Keypoints visualised on ground query 
  (bottom, coloured)
and aerial candidate 
  (top, greyscale),
colourised by RANSAC inliers
  (green)
and outliers
(red),
  with lines between correspondences. Circled regions highlight the superiority of our semantically-guided keypoints, improving consistency for cross-view re-localisation.
  }}
  \label{fig:egonn_keypoint_repeatability}
  \vspace{-5mm}
\end{figure}

While EgoNN generates repeatable keypoints in urban areas, we observe keypoint degeneracies~\cite{li2019usip} in forest environments (\figref{fig:egonn_keypoint_repeatability}). In particular, EgoNN places high confidence on keypoints in regions with low repeatability between ground and aerial viewpoints, such as the forest canopy. These are problematic as point distributions on canopy foliage can change significantly due to view-point differences, seasonal changes, growth, and weather. To prevent these degenerate keypoints, we propose a semantically-guided keypoint regression loss ($\mathcal{L}_\mathrm{seg}$) to encourage keypoint placement on repeatable features.

In the forest domain, we observe that tree trunks are one such feature with good visibility from both viewpoints, and low variation over time. We utilise a pre-trained 3D semantic segmentation model, MinkowskiNet~\cite{choy20194d}, to identify the subset of points $\mathcal{S} \subseteq \mathcal{P}$ which lie on tree trunks. Leveraging these segmentation masks, we introduce $\mathcal{L}_\mathrm{seg}$ as:
\begin{equation}
\mathcal{L}_{\mathrm{seg}} = \frac{1}{k_\mathrm{seg}}\sum_{t=1}^{M} \sum_{j=1}^{k_\mathrm{seg}} \left\|q_t-s_j\right\|_2,
    \label{eq:seg_loss}
\end{equation}
where $q_t \in \mathcal{Q}$ is the keypoint coordinate, $s_j \in \mathcal{S}$ is the coordinate of the $k_\mathrm{seg}$-nearest tree trunk points, and $\|.\|$ denotes $L_2$ norm. This loss encourages keypoints to be placed on the nearest tree trunk, with a value of $k_\mathrm{seg}$ chosen to reduce the impact of outlier segmentation labels. We add $\mathcal{L_\mathrm{seg}}$ to the total loss during training to guide keypoint selection, thus we do not require semantic point clouds during inference. \figref{fig:egonn_keypoint_repeatability} demonstrates the superiority of our approach, reducing the number of outlier keypoints placed in the canopy and improving keypoint repeatability between viewpoints.

\subsection{Ground-Aerial Factor-Graph Formulation}
\label{sec:ag_factor_graph}
In our global localisation problem, we estimate the history of robot poses $\mathcal{X} = \{\xbf_k\}_{k=1}^{n}$ with respect to an aerial reference map $\mathcal{M}$ using LiDAR and IMU measurements. We model this as a factor graph optimisation, where states are represented as $\xbf_k = [\mathbf{R}_k, \mathbf{t}_k] \in SE(3)$, with $\mathbf{R}_k \in SO(3)$ for orientation and $\mathbf{t}_k \in \mathbb{R}^3$ for translation in frame $\mathcal{M}$. 
Assuming measurements are corrupted by white Gaussian noise, the optimisation is formulated as:
\begin{equation}
    \underset{\substack{\mathcal{X}}}{\text{argmin}} \ f_\text{prior} + \sum_{\substack{k=1}}^{n} f_\text{map}(\xbf_k) + \sum_{\substack{k=1}}^{n-1} f_\text{odom}(\xbf_k,\xbf_{k+1}),
    \label{eq:optproblem}
\end{equation}
where $f_\text{prior}$ represents the prior factor, while $f_\text{odom}$ and $f_\text{map}$ correspond to odometry and map factors, respectively. The odometry factor is derived from Wildcat odometry~\cite{ramezani2022wildcat}, modelled as a non-linear least squares function linearised at the state estimates:
\begin{equation}
    \resizebox{\linewidth}{!}{
    $
    f_\text{odom} = \big\| g(\xbf^0_k,\xbf^0_{k+1}) + \mathbf{F}_k\delta\xbf_k 
    + \mathbf{G}_{k+1}\delta\xbf_{k+1} - \mathbf{T}_{k,k+1} \big\|^2_{\mathbf{\Sigma}},
    $
    }
    \label{eq:odom}
\end{equation}
where $g$ defines the odometry function, $\mathbf{F}$ and $\mathbf{G}$ are Jacobians, $\mathbf{T}_{k,k+1}$ is the relative transformation, and $\delta\xbf$ represents the state update. The notation \( \left\| \cdot \right\|^2_{\mathbf{\Sigma}} \) accounts for measurement uncertainty.

The map factor is constructed from \emph{ground-to-aerial submap registrations}, performed via GICP~\cite{segal2009generalized}. Unlike standard ICP, GICP minimises the distance between local point distributions in submaps $\mathcal{P}_G$ and $\mathcal{P}_A$:
\begin{equation}
    \underset{\substack{\mathbf{T}_{MG}}}{\text{argmin}} \ \sum_{\substack{i=1}}^m \mathbf{d}^T_i(\mathbf{\Sigma}_{Ai} + \mathbf{T}_{MG}\mathbf{\Sigma}_{Gi}\mathbf{T}_{MG}^T)^{-1}\mathbf{d}_i,
    \label{eq:gicp}
\end{equation}
where $\mathbf{d}_i$ denotes the distance from a local distribution point in $\mathcal{P}_A$ to its nearest counterpart in $\mathcal{P}_G$, after applying transformation $\mathbf{T}_{MG}$. The Hessian matrix corresponding to the robot pose~$\xbf_k$ is approximated as
\begin{equation}
    \mathbf{H} \approx  \sum_{\substack{i=1}}^m \bigg(\frac{\partial\mathbf{d}_i}{\partial\xbf_k}\bigg)^T\mathbf{\Omega}_i \bigg(\frac{\partial\mathbf{d}_i}{\partial\xbf_k}\bigg),
    \label{eq:hessian}
\end{equation}
where $\mathbf{\Omega}_i = (\mathbf{\Sigma}_{Ai} + \mathbf{T}_{MG}\mathbf{\Sigma}_{Gi}\mathbf{T}_{MG}^T)^{-1}$. The inverse of this Hessian serves as the transformation covariance $\mathbf{\Lambda}$. Upon convergence, the estimated transformation $\mathbf{T}_{MG}$ and its covariance $\mathbf{\Lambda}$ define the map factor:
\begin{equation}
    f_\text{map} = ||h(\xbf_k) + \mathbf{J}_k\delta\xbf_k - \textbf{T}_{MG}||^2_{\mathbf{\Lambda}},
    \label{eq:ref}
\end{equation}
where $h$ and $\mathbf{J}$ are the function and Jacobian matrix related to $\mathbf{T}_{MG}$. As new submaps are added, the factor graph is incrementally optimised using iSAM2~\cite{gtsam}, efficiently updating the solution without reprocessing the entire graph.
 \section{Experimental Evaluation}
\label{sec:experiment}
\noindent
\textbf{Data Collection}: To evaluate our approach, we collected data over three experiments in separate forest areas at the Queensland Centre for Advanced Technologies (QCAT), and the Venman National Park, both in Brisbane, Australia. Ground-based LiDAR data was gathered using an electric four-wheel robotic vehicle and a handheld device~\cite{de2023air}. A DJI M300 quadcopter equipped with a GPS antenna (for geo-referencing purposes) and a LiDAR system captured aerial data for the test areas.

\noindent
\textbf{Sensor Setup}: The robot and handheld device were equipped with IMU sensor and a perception pack (shown in~\figref{fig:system_overview}), featuring a Velodyne VLP-16 LiDAR sensor mounted on a servo motor inclined at $45$\degree. The sensor, rotating around the vertical axis at $0.5~Hz$, provides a $120\degree$~vertical field of view, enabling effective scanning of tree crowns. Maximum LiDAR range was set to $100~m$ with a $20~Hz$ recording rate.

\noindent
\textbf{Ground Truth}: To establish ground truth for evaluation, we designed the traversals to include loops, which allowed us to leverage an accurate SLAM system with loop-closure detection to mitigate drift and achieve precise pose estimates. We used our LiDAR SLAM system, Wildcat~\cite{ramezani2022wildcat}, configured in offline mode with loop closures enabled. This setup allowed the accurate integration of LiDAR and IMU data across the path, resulting in highly precise trajectory estimates.

\noindent
\textbf{Implementation}: Our method is implemented in C++ using the Robot Operating System (ROS). Point cloud registration with uncertainty computation was handled using GICP~\cite{kenji-fast-gicp}, while GTSAM~\cite{gtsam} was used for factor-graph optimisation. The algorithm was executed on a i7-10875H CPU (2.30GHz) unit with Ubuntu 20.04, and a single NVIDIA Quadro RTX 5000 GPU for deep re-localisation. 

\noindent
\textbf{Experimental Design:}
We evaluate our localisation method in three forest environments of increasing complexity. Experiment \datashort{} took place in a sparse forest, where a four-wheel robotic vehicle collected ground LiDAR along a 300-metre trajectory, with aerial LiDAR covering 9.5 hectares. 

Experiments \datalong{} and \emph{III} occurred in dense forests with tall undergrowth and cluttered trees, using a handheld LiDAR with IMU and GPS. Experiment \datalong{} follows a 2-kilometre loopy path over 1 hour and 15 minutes, generating a high-accuracy reference trajectory with 6.5 hectares of aerial coverage.

Experiment \datavenman{} assesses localisation in goal-directed navigation along a 1.4-kilometre open-loop path over $\sim$30 minutes, with 31 hectares of aerial coverage. Without loop closures, this scenario challenges odometry-based methods. For ground truth purposes, data was collected in two overlaid runs: one with a closed-loop segment and the other an open-loop segment, the latter used for evaluation.

\noindent
\textbf{Training Data:}
To train our semantically-guided deep re-localisation module, we utilised the V-01 and V-02 sequences from the Wild-Places dataset~\cite{knights2022wild} to form our ground queries, according to the training and testing regions proposed in the paper. These sequences were captured along a similar trajectory to our \datavenman{} experiment, but with a four-year time gap between collection. We align ground and aerial sequences following the registration protocol used in Wild-Places. Submaps from Wild-Places were sampled at a rate of $2~Hz$ along the trajectory, aggregating points captured within a one-second sliding window, and cropping submaps to a 30-metre radius of the sensor origin.

For aerial data, we used a grid-based sampling approach to generate overlapping submaps with equal coverage to ground submaps. Hence, submap centroids were sampled every 5 metres along a 2D grid covering the entire aerial map, and all points within a 30-metre horizontal radius of each centroid were kept. All submaps were then gravity-aligned and translated such that the $z$-coordinate of the lowest point in each submap begins at zero metres, and uninformative ground points were removed with a cloth simulation filter~\cite{zhangEasytoUseAirborneLiDAR2016}. To save computation, we downsampled all submaps using a 0.3-metre voxel grid filter. This produced 5222 ground submaps and 5809 aerial submaps for training.

\subsection{Global Localisation Using Deep Learning}
\noindent
\textbf{Training Setup}: To facilitate our semantically-guided keypoint regression loss, we generate semantic segmentation masks offline for all training submaps. As noted earlier, we use a MinkowskiNet~\cite{choy20194d} model pre-trained on forest environments with two classes: vegetation and trunks.
For training the re-localisation module, we follow the training protocol introduced in~\cite{komorowski2021egonn}, using a batch-hard triplet margin loss with batch size of 96 and Adam~\cite{kingmaAdamMethodStochastic2014} optimiser, trained for 120 epochs with a weight decay of $1.0e^{-4}$ and a LR of $1.0e^{-3}$, reducing the LR by a factor of 10 after epoch 80. We set $k_\mathrm{seg}$ to $5$ in our semantically-guided loss. To reduce overfitting, we adopt data augmentations including random flips, random rotations of $\pm 180^\circ$, random translation, random point jitter, and random block removal. During training, ground submaps are used as queries, and aerial submaps are potential positive and negative candidates, using positive and negative thresholds of $3~m$ and $15~m$, respectively. 

\begin{figure}[t]
  \centering
\begin{subfigure}{0.49\linewidth}
  \includegraphics[width=.99\linewidth]{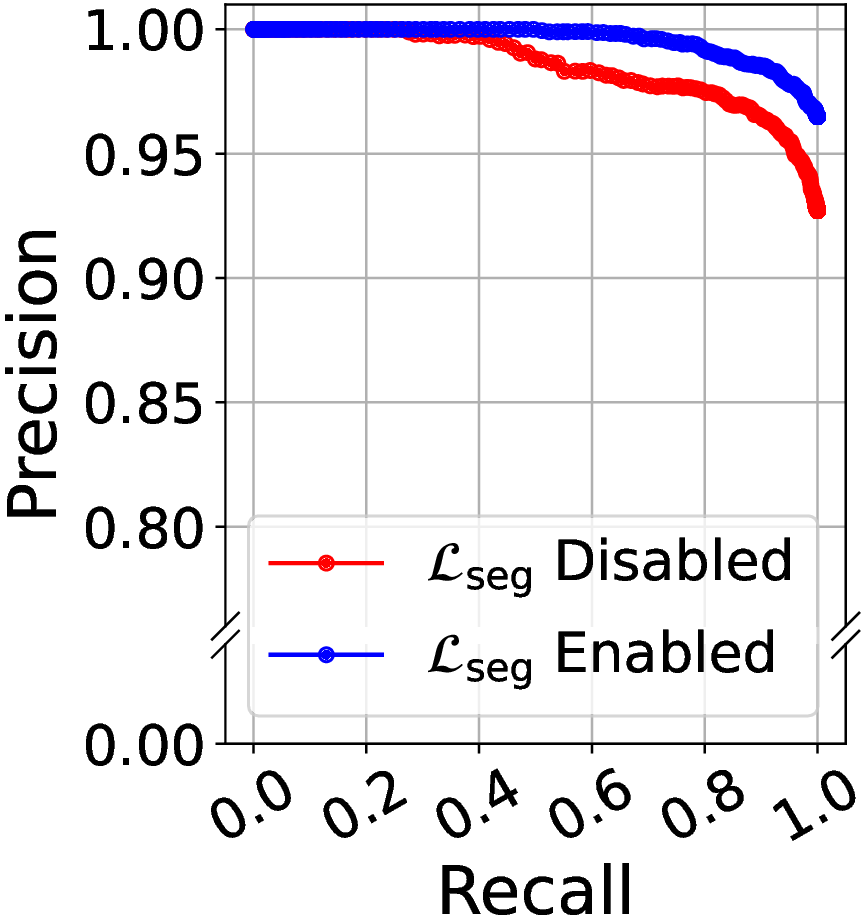}
\caption{\emph{Forest II}}
    \label{fig:egonn_pr_curve_qcat}
  \end{subfigure}
\hfill
  \begin{subfigure}{0.49\linewidth}
  \includegraphics[width=.99\linewidth]{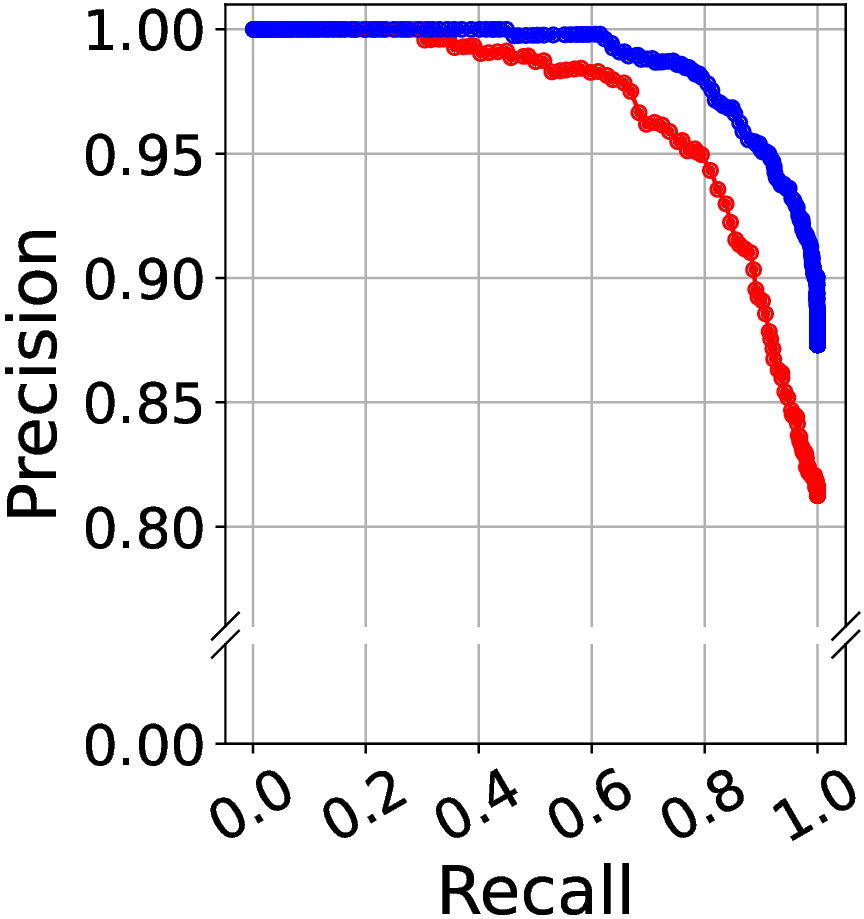}
\caption{\emph{Forest III}}
    \label{fig:egonn_pr_curve_venman}
  \end{subfigure}
  \vspace{-5mm}
  \caption{\small{Precision-Recall curves for our deep re-localisation module. We omit \datashort{} as both methods achieve $\sim$100\% $F1_{\mathrm{max}}$.
  }}
  \label{fig:egonn_pr_curves}
\end{figure}

\noindent
\textbf{Evaluation}:~We evaluate the pre-trained deep re-localisation module on the experiments introduced in the previous section, which have not been seen by the network during training. For evaluation, we pre-compute a database of aerial global descriptors for each experiment, and compute the Euclidean distance between ground query descriptors and database descriptors to find nearest candidates. We report the Precision-Recall curve in \figref{fig:egonn_pr_curves} and the corresponding maximum $F1$ score ($F1_{\mathrm{max}}$) in \tabref{tab:egonn_reloc_results}, using thresholds of $5~m$ and $20~m$ to classify true and false positives, respectively. The results demonstrate the effectiveness of our semantically-guided deep re-localisation, achieving a $F1_{\mathrm{max}}$ of $0.949$--$1.000$ across all three experiments, compared to EgoNN~\cite{komorowski2021egonn} which achieves $0.899$--$0.994$. This result suggests that enforcing consistent and repeatable keypoints improves the distinctiveness of global descriptors.

\tabref{tab:egonn_reloc_results} also shows metric localisation performance and keypoint consistency metrics. To evaluate metric localisation, we used RANSAC matching on the 128 keypoints with lowest saliency uncertainty to estimate the relative 6DoF pose, and consider initial registration successful if the relative rotation and translation error are less than $5$\degr{} and $2~m$, respectively. Note that metric localisation is only evaluated on successful top-1 retrievals that are within $20~m$ of the query, and we compute metrics on the initial registration results before applying GICP to fairly evaluate the deep re-localisation module. We report the registration success rate, and for keypoint consistency we report the inlier ratio computed by RANSAC, and keypoint repeatability, defined as $
    \frac{1}{M} \sum_{i=1}^M \mathbf{1} (\| \mathbf{R}Q_i+\mathbf{t}-\tilde{Q}_j \|_2 < \epsilon )$,
where $\mathbf{R}\in SO(3)$ and $\mathbf{t}\in\mathbb{R}^3$ are the ground truth rotation and translation between point clouds $\mathcal{P}$ and $\tilde{\mathcal{P}}$ with corresponding keypoints $Q_i\in\mathcal{Q}$ and its nearest neighbour $\tilde{Q}_j\in\tilde{\mathcal{Q}}$, $\epsilon$ is a distance threshold, and $\mathbf{1}(x)$ is an indicator function that equals 1 when $x$ is true and 0 when $x$ is false. We set $\epsilon$ to $0.5~m$ in our experiments.
\begin{table}[t!]
    \centering
    \resizebox{\linewidth}{!}{
    \begin{tabular}{lccccc}
    \hline
    Experiment & $\mathcal{L}_{\mathrm{seg}}$ & $F1_{\mathrm{max}}$ & Success Rate & Repeatability & Inlier Ratio \\ \hline
    \multirow{2}{*}{\datashort{}} & \xmark & 0.994 & 0.931 & 0.170 & 0.233 \\
     & \cmark & \textbf{1.000} & \textbf{0.989} & \textbf{0.371} & \textbf{0.439} \\ \hline
    \multirow{2}{*}{\datalong{}} & \xmark & 0.963 & 0.858 & 0.126 & 0.193 \\
     & \cmark & \textbf{0.982} & \textbf{0.969} & \textbf{0.229} & \textbf{0.299} \\ \hline
    \multirow{2}{*}{\datavenman{}} & \xmark & 0.899 & 0.893 & 0.130 & 0.208 \\
     & \cmark & \textbf{0.949} & \textbf{0.974} & \textbf{0.243} & \textbf{0.330} \\ \hline
    \end{tabular}
    }
     \caption{\small{Evaluation of our deep re-localisation module, with and without our semantically-guided keypoint regression loss ($\mathcal{L}_{\mathrm{seg}}$).}}
    \label{tab:egonn_reloc_results}
    \vspace{-5mm}
\end{table}

The results highlight the importance of our semantically-guided keypoint regression loss, achieving metric localisation success rates of $0.969$--$0.989$ on all three experiments, compared to $0.858$--$0.931$ without this loss. The effect on keypoint consistency is clear, with semantically-guided keypoints producing keypoint repeatability of $0.229$--$0.371$ and inlier ratios of $0.299$--$0.439$ across all experiments, compared to EgoNN which achieves $0.126$--$0.170$ and $0.193$--$0.233$ for repeatability and inlier ratio, respectively. The significant improvement in inlier ratio and repeatability is essential for metric localisation in this challenging cross-view scenario, where there are naturally fewer correspondences due to low overlap between point clouds from each viewpoint. A higher number of inlier correspondences improves the robustness of the RANSAC solution, thus providing a strong initial registration to prevent GICP falling into local minima.

\begin{figure}[t]
 \begin{center}
  \includegraphics[width=.85\columnwidth]{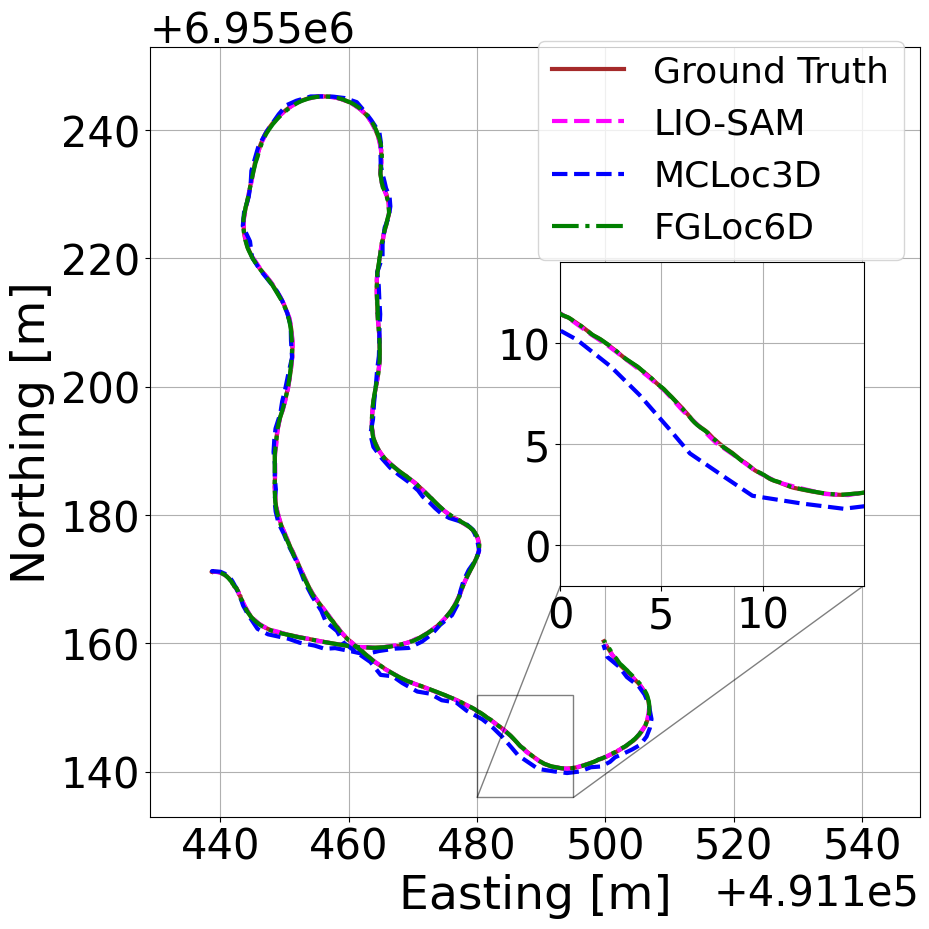}
  \\
  \includegraphics[width=.9\columnwidth]{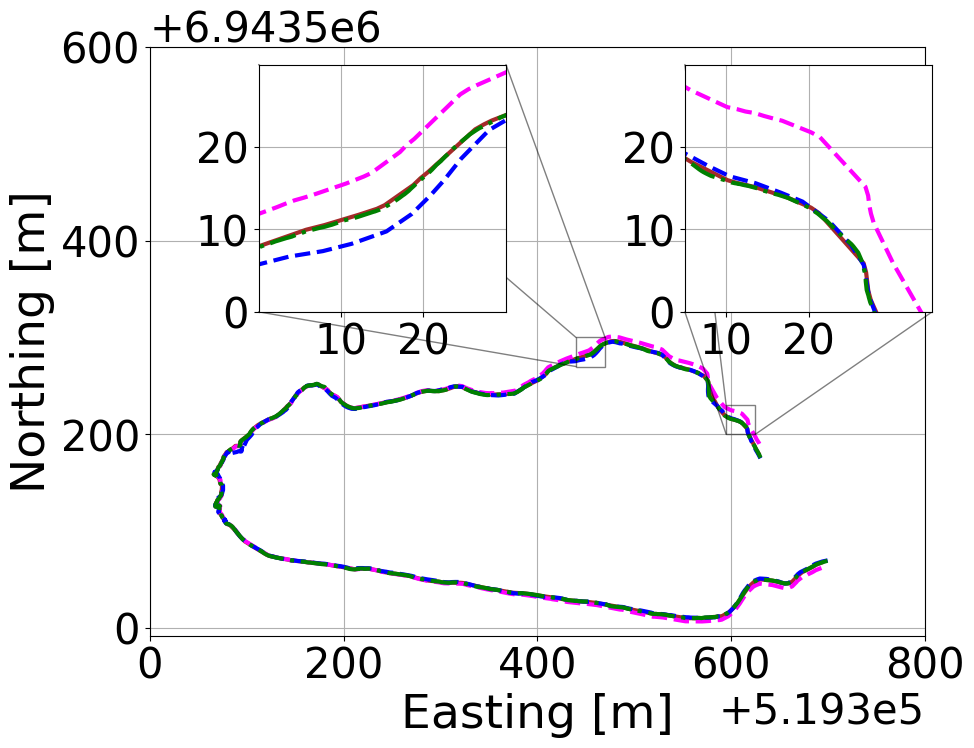}

\end{center}
\vspace{-1mm}
 \caption{\small{Trajectory comparison between our proposed localisation method FGLoc6D, ground truth (offline Wildcat SLAM~\cite{ramezani2022wildcat} with loop-closures), our previous approach MCLoc3D~\cite{de2023air}, and the LiDAR-inertial odometry LIO-SAM~\cite{shan2020lio} with loop-closure in \datashort{} (top) and \datavenman{} (bottom).}}
\label{fig:traj_comparison}
\vspace{-6mm}
 \end{figure}

\subsection{Localisation Evaluation and Comparison to Baselines}
We compare the performance of FGLoc6D with localisation baselines: MCLoc3D~\cite{de2023air}, which estimates 3DoF poses within prior aerial heightmaps using canopy map matching and Monte Carlo localisation; and LIO-SAM\cite{shan2020lio}, which performs scan-to-map LiDAR-inertial odometry with loop-closures enabled to allow accurate pose estimates. GPS performance under dense canopy is also evaluated using a baseline composed of a factor-graph optimisation integrating GPS (reference factors) with Wildcat Odometry (\ie{} LiDAR-inertial odometry without loop-closure)~\cite{ramezani2022wildcat}. Ground truth and baselines' trajectories are aligned via Umeyama using the package evo\footnote{https://github.com/MichaelGrupp/evo} for evaluation purposes.

As MCLoc3D provides poses in $SE(2)$, we first evaluate absolute translation ($x$, $y$) and heading errors ($\theta$) across all methods against ground truth for a fair comparison. We also include relative translation error in 2D to assess robustness to drift. Lastly, we compute the 6DoF absolute translation and orientation errors for all baselines, excluding MCLoc3D.

\begin{figure}
    \centering
 \includegraphics[width=0.99\columnwidth]{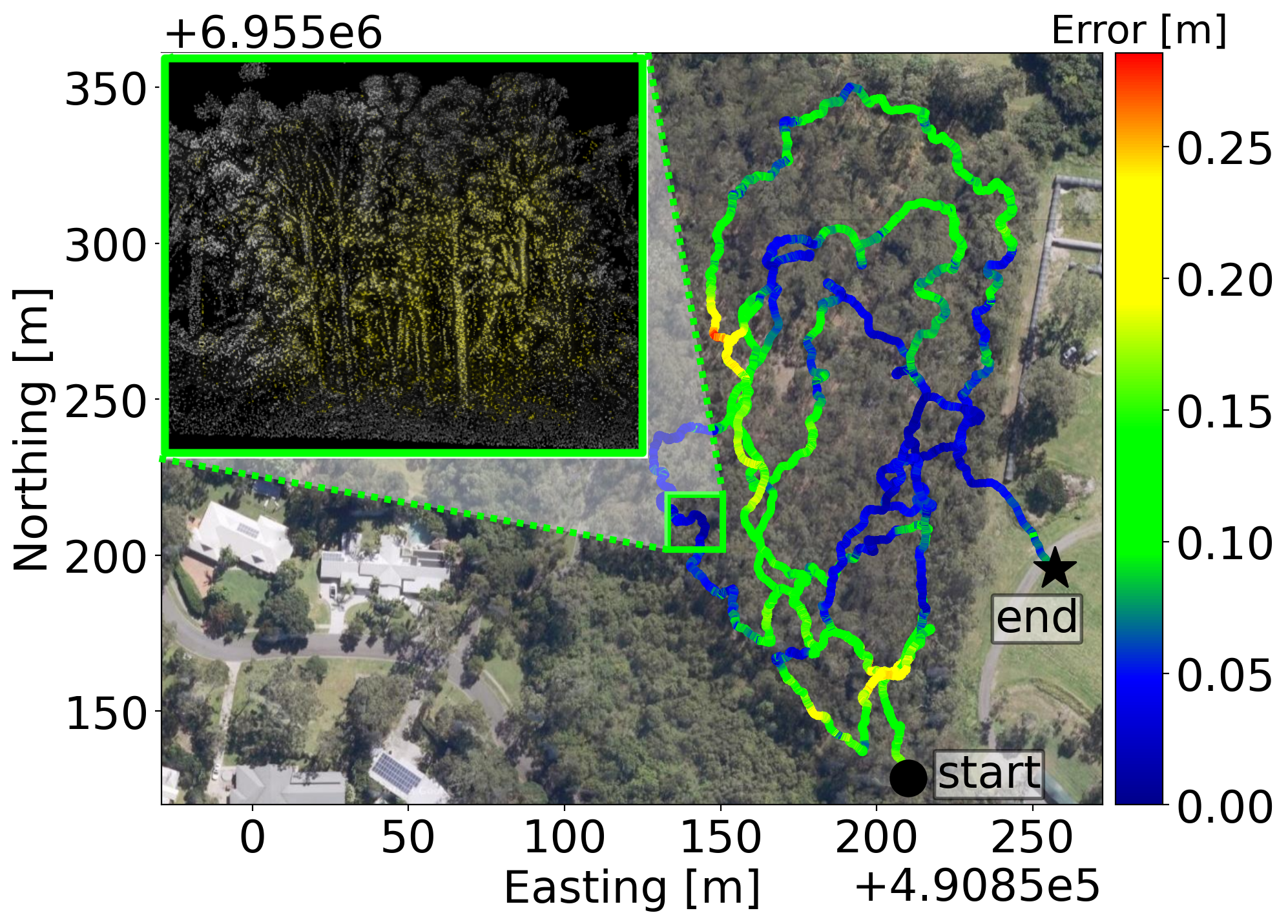}
    \caption{\small{FGLoc6D trajectory estimate in~\datalong{}. Colour code illustrates absolute translation error compared to ground truth. \textbf{Top-left}: A snapshot of ground-to-aerial registration within the area indicated by a square.}}
    \label{fig:our_satellite}
    \vspace{-2mm}
\end{figure}

\figref{fig:traj_comparison} compares the trajectories of FGLoc6D, MCLoc3D~\cite{de2023air} and LIO-SAM~\cite{shan2020lio} against ground truth in \datashort{} and \emph{III}. In both datasets, FGLoc6D closely follows the ground truth with minimal positional errors, while MCLoc3D exhibits larger deviations, as highlighted in the magnified views (top and bottom rows). In \datashort{}, \figref{fig:traj_comparison} (top), LIO-SAM accurately tracks the ground truth given the short trajectory and the incorporation of loop-closure, however, the magnified views in \datavenman{} (bottom) show high position errors due to the inherent drift of LIO-SAM. The high accuracy of our localisation in \datalong{} is further demonstrated in \figref{fig:our_satellite}, evident from the precise registration of air-ground submaps (top-left snapshot).

\begin{table}[!t]
\centering
  \resizebox{\linewidth}{!}{
  \Huge
    \begin{tabular}{l c l c c c c c} 
    \hline
         Experiment& Method &  Metric  & RMSE & median & mean & std.dev & max \\
         \hline

     \multirow{6}[6]{*}{\datashort} &\multirow{2}[2]{*}{MCLoc3D~\cite{de2023air}}    & Trans. Error [m] &  0.57 &  0.46 & 0.50 & 0.28 & 1.39\\
                                    &                               & Head. Error [deg] & 1.17 &  0.73 & 0.89 & 0.75 & 3.81 \\\cmidrule{2-8}
                                    & \multirow{2}[2]{*}{LIO-SAM~\cite{shan2020lio}}  & Trans. Error [m] & \textbf{0.05} &  \textbf{0.05} & \textbf{0.05} & \textbf{0.02} & \textbf{0.13}\\
                                    &                               & Head. Error [deg] & \textbf{0.21} &  0.13 & \textbf{0.16} & \textbf{0.13} & \textbf{0.87}\\\cmidrule{2-8}
    &\multirow{2}[2]{*}{FGLoc6D~(Ours)} & Trans. Error [m] & 0.07 &  \textbf{0.05} & 0.06 & 0.04 & 0.19 \\
     &                                         & Head. Error [deg] & 0.25 &  \textbf{0.11} & \textbf{0.16} & 0.19 & 1.10  \\
                               \hline

    \multirow{6}[6]{*}{\datalong} & \multirow{2}[2]{*}{MCLoc3D~\cite{de2023air}} & Trans. Error [m] &  0.50 &  0.41 & 0.44 & 0.24 & 1.31 \\
                                 &                              & Head. Error [deg] & 2.55 &  1.35 & 1.73 & 1.87 & 31.96 \\\cmidrule{2-8}
                            & \multirow{2}[2]{*}{LIO-SAM~\cite{shan2020lio}}        & Trans. Error [m] &  0.19 &  0.15 & 0.17 & 0.09 & 0.80\\
                                                &                & Head. Error [deg] & 1.10 &  0.65 & 0.83 & 0.71 & 8.25 \\\cmidrule{2-8}
                 &\multirow{2}[2]{*}{FGLoc6D~(Ours)} & Trans. Error [m] & \textbf{0.11} &  \textbf{0.08} & \textbf{0.09} & \textbf{0.06} & \textbf{0.29}\\
                                            &               & Head. Error [deg] & \textbf{0.70} &  \textbf{0.28} & \textbf{0.47} & \textbf{0.52} & \textbf{6.05} \\
                 \hline
       \multirow{6}[6]{*}{\datavenman} & \multirow{2}[2]{*}{MCLoc3D~\cite{de2023air}} & Trans. Error [m] & 1.40 &  0.74 & 0.99 & 0.99 & 8.16 \\
                 &                                                   & Head. Error [deg] &  3.16 &  1.22 & 1.99 & 2.45 & 20.94 \\\cmidrule{2-8}
                 & \multirow{2}[2]{*}{LIO-SAM~\cite{shan2020lio}}        & Trans. Error [m] & 3.40 &  0.94 & 2.43 & 2.38 & 8.77\\
                                     &                & Head. Error [deg] &  1.34 &  0.93 & 1.08 & 0.79 & 4.25 \\\cmidrule{2-8}
                 &\multirow{2}[2]{*}{FGLoc6D~(Ours)} & Trans. Error [m] & \textbf{0.21} &  \textbf{0.19} & \textbf{0.19} & \textbf{0.10} & \textbf{0.69} \\
                                            &               & Head. Error [deg] & \textbf{0.56} &  \textbf{0.29} & \textbf{0.40} & \textbf{0.39} & \textbf{2.97}\\
                 \hline
    \end{tabular}
    }
\caption{\small{3DoF pose error evaluation in \datashort{}, \emph{II} and \emph{III} comparing FGLoc6D (ours) with MCLoc3D and LIO-SAM baselines. Trans. Error indicates absolute translation ($x,y$) errors, and Head. Error shows absolute heading ($\theta$) errors, both relative to the ground truth trajectory. Bold numbers indicate the lowest values.}}
\label{table:3dof_errors}
\vspace{-5mm}
\end{table}

As observed from the absolute translation and heading errors (\tabref{table:3dof_errors}), in \datashort{}, FGLoc6D performs on par with LIO-SAM which benefited from the short trajectory (because of scan-to-map matching) and loop-closure incorporation. In  \datalong{} and \emph{III}, FGLoc6D outperforms all baselines, providing more accurate pose estimates and significantly improving over MCLoc3D. These results are validated by the 6DoF pose error statistics (\tabref{table:6dof_errors}), where FGLoc6D achieves the highest accuracy among all baselines, maintaining absolute position and orientation RMSE under $0.25~m$ and $0.93$\degr{} in the challenging datasets \datalong{} and \emph{III}. 

While LIO-SAM with loop-closure achieves low translation and orientation errors in \datashort{} and \emph{II}, its performance degrades significantly in \datavenman{}, with maximum translation errors exceeding $8~m$ (\tabref{table:3dof_errors}) and $10~m$ (\tabref{table:6dof_errors}), highlighting the impact of error accumulation. This experiment demonstrates that errors in SLAM or odometry methods grow unbounded in open-loop paths (\datavenman{}), which can critically affect mission success in robotic downstream tasks. In contrast, FGLoc6D’s incorporation of graph optimisation with aerial map and LiDAR odometry factors ensures constant drift-free and accurate pose estimates. 

The GPS solution exhibits high RMSE, $5.77~m$ and $4.91$\degr{}, and a large maximum position error ($12.36~m$) in \datavenman{} (see \tabref{table:6dof_errors}), confirming that degraded GPS signals are unreliable for accurate position estimation beneath the canopy, even when fused with LiDAR-inertial odometry.

\begin{table}[!t]
\centering
  \resizebox{\linewidth}{!}{
  \Huge
    \begin{tabular}{l c l c c c c c} 
    \hline
         Experiment& Method &  Metric  & RMSE & median & mean & std.dev & max \\
         \hline
                  \multirow{4}[4]{*}{\datashort} & \multirow{2}[2]{*}{LIO-SAM~\cite{shan2020lio}}       & Trans. Error [m] & \textbf{0.06} &  \textbf{0.05} & \textbf{0.05} & \textbf{0.02} & \textbf{0.13} \\
                  &               & Rot. Error [deg] & \textbf{0.30} &  \textbf{0.25} & \textbf{0.26} & \textbf{0.14} & \textbf{0.94} \\\cmidrule{2-8}
                  &\multirow{2}[2]{*}{FGLoc6D~(Ours)} & Trans. Error [m] &  0.08 &  \textbf{0.05} & 0.06 & 0.04 & 0.19 \\
                  &               & Rot. Error [deg] & 0.34 &  0.26 & 0.29 & 0.18 & 1.12 \\
                               \hline
     \multirow{6}[6]{*}{\datalong} &\multirow{2}[2]{*}{GPS+LiDAR Odom} & Trans. Error [m] & 2.96 &  2.44 & 2.65 & 1.32 & 8.15\\
                  &               & Rot. Error [deg] & 5.44 &  3.96 & 4.78 & 2.59 & 11.75\\\cmidrule{2-8}
                  & \multirow{2}[2]{*}{LIO-SAM~\cite{shan2020lio}}       & Trans. Error [m] & 0.33 &  0.27 & 0.30 & 0.14 & 1.01\\
                  &               & Rot. Error [deg] & 1.53 &  1.16 & 1.30 & 0.79 & 11.18\\\cmidrule{2-8}
                  &\multirow{2}[2]{*}{FGLoc6D~(Ours)} & Trans. Error [m] & \textbf{0.12} &  \textbf{0.10} & \textbf{0.10} & \textbf{0.06} & \textbf{0.29}\\
                  &               & Rot. Error [deg] & \textbf{0.93} &  \textbf{0.56} & \textbf{0.73} & \textbf{0.57} & \textbf{6.09}\\
                               \hline

    \multirow{6}[6]{*}{\datavenman} & \multirow{2}[2]{*}{GPS+Lidar Odom} & Trans. Error [m] & 5.77 &  4.87 & 5.32 & 2.25 & 12.36\\
                 &                & Rot. Error [deg] & 4.91 &  4.64 & 4.75 & 1.25 & 7.95\\\cmidrule{2-8}
                 & \multirow{2}[2]{*}{LIO-SAM~\cite{shan2020lio}}        & Trans. Error [m] & 3.60 &  1.10 & 2.60 & 2.49 & 10.71\\
                 &                & Rot. Error [deg] & 1.93 &  1.26 & 1.61 & 1.07 & 4.93\\\cmidrule{2-8}
                 &\multirow{2}[2]{*}{FGLoc6D~(Ours)} & Trans. Error [m] & \textbf{0.25} &  \textbf{0.22} & \textbf{0.23} & \textbf{0.09} & \textbf{0.69}\\
                 &               & Rot. Error [deg] & \textbf{0.86} &  \textbf{0.58} & \textbf{0.71} & \textbf{0.49} & \textbf{3.34}\\
                 \hline
    \end{tabular}
    }
\caption{\small{Evaluation of 6DoF pose errors in \datashort{}, \emph{II} and \emph{III}, comparing FGLoc6D with GPS-based factor graph localisation and LIO-SAM baselines. Trans. Error represents absolute translation errors while Rot. Error denotes absolute rotation errors, both relative to ground truth. GPS measurements were not recorded in \datashort, so values are not displayed on the table. Bold numbers indicate the lowest errors.}}
\label{table:6dof_errors}
\vspace{-5mm}
\end{table}

\begin{figure}[ht!]
 \begin{center}
  \includegraphics[width=.9\columnwidth]{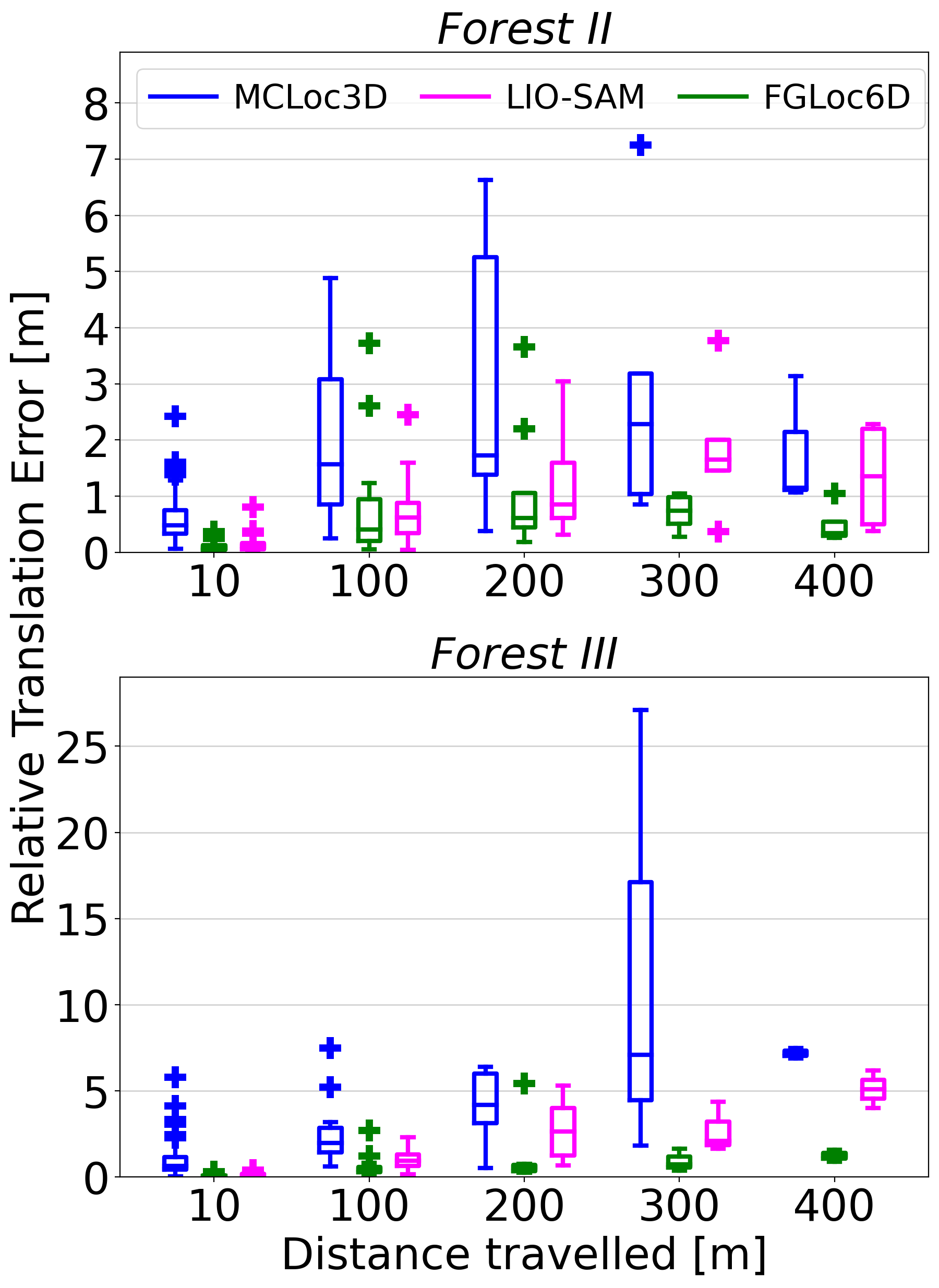}\
\end{center}
\vspace{-2mm}
 \caption{\small{Comparison of the relative translation errors of distinct pose estimation methods with respect to the ground truth in datasets \datalong{} (top) and \datavenman{} (bottom).}}
\label{fig:rpe_experiments_box}
\end{figure}

\figref{fig:rpe_experiments_box} presents the relative translation errors (RTE) of all localisation approaches against the ground truth. In \datalong{} (top) and \datavenman{} (bottom), FGLoc6D clearly outperforms baselines, maintaining most RTE values below $1~m$ and $1.5~m$, respectively, across all distance intervals. MCLoc6D exhibits higher RTE in both datasets, likely due to dense canopy areas affecting the map matching and pose estimation, accordingly. The elevated RTEs ($>1.5~m$) of LIO-SAM in \datalong{} (top) highlight that the challenging irregular forest structure can slightly impact the overall position estimation in LiDAR SLAM, despite the incorporation of loop-closures. The error accumulation of LIO-SAM becomes even more evident in \datavenman{}, \figref{fig:rpe_experiments_box} (bottom), where the RTE increases steeply, reaching $5~m$. The consistently low RTE of FGLoc6D in both datasets demonstrates that our proposed method remains drift-free in all tested environments regardless of robot trajectory.          

\subsection{Runtime Analysis}
To demonstrate that our presented system can run online, we evaluate the computation time for each component. The timing results are collected by running the pre-trained deep learning models on GPU, and the rest of the pipeline on CPU as described in \secref{sec:experiment}.~\tabref{table:runtime} reports a breakdown of individual modules’ runtime in our pipeline. The total runtime (for the \datalong{} experiment) is less than a second, allowing the system to run online. Once re-localisation is performed, the low-level odometry and back-end optimisation altogether can run at $\sim 3~Hz$.

\begin{table}[t]
\centering
  \resizebox{\linewidth}{!}{
  \Huge
    \begin{tabular}{c c c c c} 
    \hline
Runtime(s) & Odometry & Registration & FG Optimisation & Re-Localisation \\
    \hline
mean & 0.055  & 0.288 & 0.038 & 0.418 \\ 
std.dev & $\pm$0.012 & $\pm$0.063 & $\pm$0.024 & $\pm$0.189 \\
     \hline
    \end{tabular}
    }
\caption{\small{Runtime analysis of each component in our global localisation framework.}}
\label{table:runtime}
\vspace{-5mm}
\end{table}

 \section{Conclusion and Future Work}

We introduced a 6DoF pose estimation framework that integrates cross-view factor graph optimisation with a lightweight semantically-guided deep re-localisation technique, enabling precise global localisation of a ground (below canopy) robot against an aerial (above canopy) map in forest environments. Extensive experiments in challenging forest conditions demonstrate our method’s superior accuracy and robustness compared to existing baselines. A key technical achievement is the formulation of the localisation problem as a bipartite graph, combining ground-to-aerial unary factors with model-based and data-driven models for global optimisation. Additionally, the use of GICP enhances point cloud registration by reducing distributional differences between ground and aerial submaps, thus improving the precision of robot pose estimates.
The real-time capability of the system has been validated through runtime analysis, illustrating that the presented pipeline can operate efficiently under practical conditions.
In future work we will explore transformer-based re-localisation approaches for greater generalisability across varying forest environments and the use of GPU acceleration for faster registration and more scalable computation.

\section*{Acknowledgments}
The authors thank CSIRO Robotics members for their hardware and software support. This work was partially supported by the SILVANUS Project through European Commission Funding on the Horizon 2020 under Grant H2020-LC-GD-2020 and Grant 101037247.

\balance{}

\bibliographystyle{IEEEtran}
        \bibliography{ref}

\end{document}